\documentclass[11pt,a4paper]{article}
\usepackage{authblk}
\usepackage[hyperref]{acl2018}
\usepackage{times}
\usepackage{latexsym}
\usepackage{multirow}
\usepackage{hhline}
\usepackage{booktabs}
\usepackage{enumitem}
\usepackage{graphicx}
\graphicspath{ {figures/} }
\usepackage{footnote}

\usepackage{url}

\aclfinalcopy 

\newcommand{\specialcell}[2][c]{%
  \begin{tabular}[#1]{@{}c@{}}#2\end{tabular}}

\title{Breaking NLI Systems\\with Sentences that Require Simple Lexical Inferences}

\author[1]{\bf Max Glockner}
\author[2]{\bf Vered Shwartz}
\author[2]{\bf Yoav Goldberg}
\affil[1]{Computer Science Department, TU Darmstadt, Germany}
\affil[2]{Computer Science Department, Bar-Ilan University, Ramat-Gan, Israel}
\affil[ ]{\tt ~\{maxg216,vered1986,yoav.goldberg\}@gmail.com}

\date{}

\interfootnotelinepenalty=10000{}

\begin{document}
\maketitle{}

\vspace*{-25pt}
\begin{abstract}
We create a new NLI test set that shows the deficiency of state-of-the-art models in inferences that require lexical and world knowledge. 
The new examples are simpler than the SNLI test set, containing sentences that differ by at most one word from sentences in the training set. 
Yet, the performance on the new test set is substantially worse across systems trained on SNLI, demonstrating that these systems are limited in their generalization ability, failing to capture many simple inferences.
\end{abstract}

\vspace*{-10pt}
\section{Introduction}
\label{sec:intro}
Recognizing textual entailment (RTE) \cite{dagan2013recognizing}, recently framed as natural language inference (NLI) \cite{D15-1075} is a task concerned with identifying whether a 
\emph{premise} sentence entails, contradicts or is neutral with the \emph{hypothesis} sentence. Following the release of the large-scale SNLI dataset \cite{D15-1075}, many end-to-end neural models have been developed for the task, achieving high accuracy on the test set. As opposed to previous-generation methods, which relied heavily on lexical resources, neural models only make use of pre-trained word embeddings. The few efforts to incorporate external lexical knowledge resulted in negligible performance gain \cite{chen-EtAl:2017b:natural}. This raises the question whether (1) neural methods are inherently stronger, obviating the need of external lexical knowledge; (2) large-scale training data allows for implicit learning of previously explicit lexical knowledge; or (3) the NLI datasets are simpler than early RTE datasets, requiring less knowledge. 

\begin{table}[ht]
\small
\center
	\begin{tabular}{lc}
\toprule
	\textbf{Premise/Hypothesis} & \textbf{Label} \\  \midrule
		The man is holding a saxophone & \multirow{2}{*}{contradiction\footnotemark} \\
        The man is holding an electric guitar & \\
        \midrule
		A little girl is very sad. & \multirow{2}{*}{entailment} \\ 
        A little girl is very unhappy.&  \\
        \midrule
		A couple drinking wine & \multirow{2}{*}{neutral} \\
        A couple drinking champagne &  \\
    \bottomrule
  \end{tabular}
    \vspace*{-7pt}
	\caption{Examples from the new test set.} 
    \label{tab:data_examples}
    \vspace*{-15pt}
\end{table}
\footnotetext{The contradiction example follows the assumption in \newcite{D15-1075} that the premise contains the most prominent information in the event, hence the premise can't describe the event of a man holding both instruments.}

In this paper we show that state-of-the-art NLI systems are limited in their generalization ability, and fail to capture many simple inferences that require lexical and world knowledge. Inspired by the work of \newcite{jia-liang:2017:EMNLP2017} on reading comprehension, we create a new NLI test set with examples that capture various kinds of lexical knowledge (Table~\ref{tab:data_examples}). For example, that \textit{champagne} is a type of \textit{wine} (hypernymy), and that \textit{saxophone} and \textit{electric guitar} are different musical instruments (co-hyponyms). To isolate lexical knowledge aspects, our constructed examples contain only words that appear both in the training set and in pre-trained embeddings, and differ by a single word from sentences in the training set.

The performance on the new test set is substantially worse across systems, demonstrating that the SNLI test set alone is not a sufficient measure of language understanding capabilities. Our results are in line with \newcite{gururangan2018annotation} and \newcite{hypothesis-only-baselines-in-natural-language-inference}, who showed that the label can be identified by looking only at the hypothesis and exploiting annotation artifacts such as word choice and sentence length. 

Further investigation shows that what mostly affects the systems' ability to correctly predict a test example is the amount of similar examples found in the training set. Given that training data will always be limited, this is a rather inefficient way to learn lexical inferences, stressing the need to develop methods that do this more effectively. Our test set can be used to evaluate such models' ability to recognize lexical inferences, and it is available at \href{https://github.com/BIU-NLP/Breaking_NLI}{https://github.com/} \href{https://github.com/BIU-NLP/Breaking_NLI/}{BIU-NLP/Breaking\_NLI}.

\section{Background}
\label{sec:background}
\paragraph{NLI Datasets.} The SNLI dataset \cite[Stanford Natural Language Inference,][]{D15-1075} consists of 570k sentence-pairs manually labeled as entailment, contradiction, and neutral. Premises are image captions from \newcite{young2014image}, while hypotheses were generated by crowd-sourced workers who were shown a premise and asked to generate entailing, contradicting, and neutral sentences. Workers were instructed to judge the relation between sentences \emph{given that they describe the same event}. Hence, sentences that differ by a single mutually-exclusive term should be considered contradicting, as in ``The president visited Alabama'' and ``The president visited Mississippi''. This differs from traditional RTE datasets, which do not assume event coreference, and in which such sentence-pairs would be considered neutral. 

Following criticism on the simplicity of the dataset, stemming mostly from its narrow domain, two additional datasets have been collected. The MultiNLI  dataset \cite[Multi-Genre Natural Language Inference,][]{williams2017broad} was collected similarly to SNLI, though covering a wider range of genres, and supporting a cross-genre evaluation. 
The SciTail dataset \cite{scitail}, created from science exams, is somewhat different from the two datasets, being smaller (27,026 examples), and labeled only as entailment or neutral. The domain makes this dataset different in nature from the other two datasets, and it consists of more factual sentences rather than scene descriptions.

\paragraph{Neural Approaches for NLI.} Following the release of SNLI, there has been tremendous interest in the task, and many end-to-end neural models were developed, achieving promising results.\footnote{See the SNLI leaderboard for a comprehensive list: {\url{https://nlp.stanford.edu/projects/snli/}}.} 
Methods are divided into two main approaches. Sentence-encoding models \cite[e.g.][]{D15-1075,bowman2016fast,nie2017shortcut,shen2018reinforced} encode the premise and hypothesis individually, while attention-based models align words in the premise with similar words in the hypothesis, encoding the two sentences together \cite[e.g.][]{rocktaschel2016reasoning,chen-EtAl:2017a:Long3}.

\paragraph{External Lexical Knowledge.} Traditional RTE methods typically relied on resources such as WordNet \cite{fellbaum1998wordnet} to identify lexical inferences. Conversely, neural methods rely solely on pre-trained word embeddings, yet, they achieve high accuracy on SNLI. 

The only neural model to date that incorporates external lexical knowledge (from WordNet) is KIM \cite{chen-EtAl:2017b:natural}, however, gaining only a small addition of 0.6 points in accuracy on the SNLI test set. This raises the question whether the small performance gap is a result of the model not capturing lexical knowledge well, or the SNLI test set not requiring this knowledge in the first place.

\section{Data Collection}
\label{sec:data_collection}
We construct a test set with the goal of evaluating the ability of state-of-the-art NLI models to make inferences that require simple lexical knowledge. We automatically generate sentence pairs (\S\ref{sec:generating_examples}) which are then manually verified (\S\ref{sec:annotation}).

\subsection{Generating Adversarial Examples}
\label{sec:generating_examples}

In order to isolate the lexical knowledge aspects, the premises are taken from the SNLI training set. For each premise we generate several hypotheses by replacing a single word within the premise by a different word. We also allow some multi-word noun phrases (``electric guitar'') and adapt determiners and prepositions when needed. 

We focus on generating only \textit{entailment} and \textit{contradiction} examples, while \textit{neutral} examples may be generated as a by-product. \textit{Entailment} examples are generated by replacing a word with its synonym or hypernym, while \textit{contradiction} examples are created by replacing words with mutually exclusive co-hyponyms and antonyms (see Table~\ref{tab:data_examples}). The generation steps are detailed below.

\paragraph{Replacement Words.} We collected the replacement words using online resources for English learning.\footnote{\scriptsize\url{www.enchantedlearning.com}, \url{www.smart-words.org}} The newly introduced words are all present in the SNLI training set: from occurrence in a single training example (``Portugal'') up to 248,051 examples (``man''), with a mean of 3,663.1 and a median of 149.5. The words are also available in the pre-trained embeddings vocabulary. The goal of this constraint is to isolate lexical knowledge aspects, and evaluate the models' ability to generalize and make new inferences for known words. 

Replacement words are divided into topical categories detailed in Table~\ref{tab:accuracy_by_category}. In several categories we applied additional processing to ensure that examples are indeed mutually-exclusive, topically-similar, and interchangeable in context. We included WordNet antonyms with the same part-of-speech and with a cosine similarity score above a threshold, using GloVe \cite{pennington-socher-manning:2014:EMNLP2014}. In \textit{nationalities} and \textit{countries} we focused on countries which are related geographically \textit{(Japan, China)} or culturally \textit{(Argentina, Spain)}.

\begin{table}[t]
\small
\center
	  \begin{tabular}{l r r}
\toprule
	& \textbf{SNLI Test} & \textbf{New Test} \\  \midrule
		\multicolumn{2}{l}{\textbf{Instances:}}\\
		\textit{contradiction} & 3,236 & 7,164 \\
		\textit{entailment} & 3,364 & 982 \\
		\textit{neutral} & 3,215 & 47 \\
		Overall & 9,815 & 8,193 \\
		\midrule
		\multicolumn{2}{l}{\textbf{Fleiss $\kappa$:}}\\
		\textit{contradiction} & 0.77 & 0.61 \\
		\textit{entailment} & 0.69 & 0.90 \\
		Overall & 0.67 & 0.61\\
        \midrule
        \multicolumn{2}{l}{\textbf{Estimated human performance:}}\\
        & 87.7\% & 94.1\% \\
    \bottomrule
  \end{tabular}
    \vspace*{-7pt}
	\caption{Statistics of the test sets. 9,815 is the number of samples with majority agreement in the  SNLI test set, whose full size is 9,824.} 
    \label{tab:agreement}
    \vspace*{-12pt}
\end{table}

\begin{table*}[t]
\small
\centering
	\begin{tabular}{c c c c c c}
\toprule
\textbf{Model} & \textbf{Train set} & \textbf{SNLI test set} & \textbf{New test set} & $\Delta$ \\ 
\midrule
\multirow{3}{*}{\specialcell{Decomposable Attention \\ \cite{parikh-EtAl:2016:EMNLP2016}}} & SNLI & 84.7\% & 51.9\% & -32.8 \\ 
& MultiNLI + SNLI & 84.9\% & 65.8\% & -19.1 \\ 
& SciTail + SNLI & 85.0\% & 49.0\% & -36.0 \\ 
\midrule
\multirow{3}{*}{ESIM \cite{chen-EtAl:2017a:Long3}} & SNLI & 87.9\% & 65.6\%& -22.3\\ 
& MultiNLI + SNLI & 86.3\% &74.9\% & -11.4\\
& SciTail + SNLI & 88.3\% & 67.7\%& -20.6\\ 
\midrule
\multirow{3}{*}{\specialcell{Residual-Stacked-Encoder \\ \cite{nie2017shortcut}}} & SNLI & 86.0\% & 62.2\% & -23.8\\ 
& MultiNLI + SNLI & 84.6\% & 68.2\%& -16.8\\ 
& SciTail + SNLI & 85.0\% & 60.1\%& -24.9\\ 
\midrule
WordNet Baseline & - & - & 85.8\% & - \\ 
KIM \cite{chen-EtAl:2017b:natural} & SNLI & 88.6\% & 83.5\% & -5.1 \\ 
\bottomrule
\end{tabular}

\vspace*{-7pt}
\caption{Accuracy of various models trained on SNLI or a union of SNLI with another dataset (MultiNLI, SciTail), and tested on the original SNLI test set and the new test set.}
\label{tab:results}
\vspace*{-12pt}
\end{table*} 

\paragraph{Sentence-Pairs.} To avoid introducing new information not present in the training data, we sampled premises from the SNLI training set that contain words from our lists, and generated hypotheses by replacing the selected word with its replacement. Some of the generated sentences may be ungrammatical or nonsensical, for instance, when replacing \textit{Jordan} with \textit{Syria} in sentences discussing \textit{Michael Jordan}. We used Wikipedia bigrams\footnote{\scriptsize\url{github.com/rmaestre/Wikipedia-Bigram-Open-Datasets}} to discard sentences in which the replaced word created a bigram with less than 10 occurrences. 

\subsection{Manual Verification}
\label{sec:annotation}

We manually verify the correctness of the automatically constructed examples using crowdsourced workers in Amazon Mechanical Turk. To ensure the quality of workers, we applied a qualification test and required a 99\% approval rate for at least 1,000 prior tasks. We assigned each annotation to 3 workers.

Following the SNLI guidelines, we instructed the workers to consider the sentences as describing the same event, but we simplified the annotation process into answering 3 simple yes/no questions:

\vspace{-5pt}
\begin{enumerate}[noitemsep,leftmargin=*] 
	\setlength{\itemsep}{1pt}
	\setlength{\parskip}{0pt}
	\setlength{\parsep}{0pt}
  \item Do the sentences describe the same event?
  \item Does the new sentence (hypothesis) add new information to the original sentence (premise)?
  \item Is the new sentence incorrect/ungrammatical?
\end{enumerate}
\vspace{-5pt}

We then discarded any sentence-pair in which at least one worker answered the third question positively. If the answer to the first question was negative, we considered the label as \textit{contradiction}. Otherwise, we considered the label as \textit{entailment} if the answer to the second question was negative and \textit{neutral} if it was positive. We used the majority vote to determine the gold label. 

The annotations yielded substantial agreement, with Fleiss' Kappa $\kappa = 0.61$ \cite{landis1977measurement}. We estimate human performance to 94.1\%, using the method described in \newcite{gong2017natural}, showing that the new test set is substantially easier to humans than SNLI. Table~\ref{tab:agreement} provides additional statistics on the test set.\footnote{We note that due to its bias towards \textit{contradiction}, the new test set can neither be used for training, nor serve as a main evaluation set for NLI. Instead, we suggest to use it in addition to the original test set in order to test a model's ability to handle lexical inferences.} 

\section{Evaluation}
\label{sec:evaluation}
\subsection{Models}
\label{sec:models}

\paragraph{Without External Knowledge.} We chose 3 representative models in different approaches (sentence encoding and/or attention): \textsc{Residual-Stacked-Encoder} \cite{nie2017shortcut} is a biLSTM-based single sentence-encoding model without attention. As opposed to traditional multi-layer biLSTMs, the input to each next layer is the concatenation of the word embedding and the summation of outputs from previous layers. \textsc{Esim} \cite[Enhanced Sequential Inference Model, ][]{chen-EtAl:2017a:Long3} is a hybrid TreeLSTM-based and biLSTM-based model. We use the biLSTM model, which uses an inter-sentence attention mechanism to align words across sentences. Finally, \textsc{Decomposable Attention} \cite{parikh-EtAl:2016:EMNLP2016} performs soft alignment of words from the premise to words in the hypothesis using attention mechanism, and decomposes the task into comparison of aligned words. Lexical-level decisions are merged to produce the final classification. We use the AllenNLP re-implementation,\footnote{\url{http://allennlp.org/models}} which does not implement the optional intra-sentence attention, and achieves an accuracy of 84.7\% on the SNLI test set, comparable to 86.3\% by the original system.

We chose models which are amongst the best performing within their approaches (excluding ensembles) and have available code. All models are based on pre-trained GloVe embeddings \cite{pennington-socher-manning:2014:EMNLP2014}, which are either fine-tuned during training (\textsc{Residual-Stacked-Encoder} and \textsc{esim}) or stay fixed (\textsc{Decomposable Attention}). All models predict the label using a concatenation of features derived from the sentence representations (e.g. maximum, mean), for example as in \newcite{mou2016natural}. We use the recommended hyper-parameters for each model, as they appear in the provided code. 

\paragraph{With External Knowledge.} We provide a simple \textsc{WordNet baseline}, in which we classify a sentence-pair according to the WordNet relation that holds between the original word $w_p$ and the replaced word $w_h$. We predict \textit{entailment} if $w_p$ is a hyponym of $w_h$ or if they are synonyms, \textit{neutral} if $w_p$ is a hypernym of $w_h$, and \textit{contradiction} if $w_p$ and $w_h$ are antonyms or if they share a common hypernym ancestor (up to 2 edges). Word pairs with no WordNet relations are classified as \textit{other}. 

We also report the performance of \textsc{kim} \cite[Knowledge-based Inference Model, ][]{chen-EtAl:2017b:natural}, an extension of \textsc{esim} with external knowledge from WordNet, which was kindly provided to us by Qian Chen. \textsc{Kim} improves the attention mechanism by taking into account the existence of WordNet relations between the words. The lexical inference component, operating over pairs of aligned words, is enriched with a vector encoding the specific WordNet relations between the words. 

\begin{table*}[t]
\hspace*{-15pt}
\small
\centering
\begin{tabular}{l | l r c c c c c c} 
\toprule
\specialcell{\scriptsize\textbf{Dominant}\\\scriptsize\textbf{Label}} & \textbf{Category} & \textbf{Instances} & \specialcell{\textbf{Example}\\\textbf{Words}} & \specialcell{\textbf{Decomposable}\\\textbf{Attention}} & \textbf{ESIM} & \specialcell{\textbf{Residual}\\\textbf{Encoders}} & \specialcell{\textbf{WordNet}\\\textbf{Baseline}} & \textbf{KIM}  \\ \midrule
\multirow{12}{*}{Cont.} & antonyms & 1,147 & \textit{loves - dislikes} & 41.6\% & 70.4\%& 58.2\% &  95.5\% & 86.5\% \\ 
& cardinals & 759 &\textit{five - seven} & 53.5\% &75.5\% & 53.1\% & 98.6\% & 93.4\% \\ 
& nationalities & 755 & \textit{Greek - Italian} & 37.5\% &35.9\% & 70.9\% & 78.5\% & 73.5\% \\ 
& drinks & 731 & \textit{lemonade - beer} & 52.9\% &63.7\% & 52.0\% & 94.8\% & 96.6\% \\ 
& antonyms (WN) & 706 & \textit{sitting - standing} & 55.1\% & 74.6\%& 67.9\% & 94.5\% & 78.8\% \\
& colors & 699 & \textit{red - blue} & 85.0\% &96.1\% & 87.0\% & 98.7\% & 98.3\% \\ 
& ordinals & 663 & \textit{fifth - 16th} & 2.1\% &21.0\% & 5.4\% &40.7\% & 56.6\% \\ 
& countries & 613 & \textit{Mexico - Peru} & 15.2\% &25.4\% & 66.2\% & 100.0\% & 70.8\% \\ 
& rooms & 595 & \textit{kitchen - bathroom} & 59.2\% &69.4\% & 63.4\% & 89.9\% & 77.6\% \\ 
& materials & 397 & \textit{stone - glass} & 65.2\% & 89.7\%& 79.9\% & 75.3\% & 98.7\% \\
& vegetables & 109 & \textit{tomato -potato} & 43.1\% &31.2\% & 37.6\% & 86.2\% & 79.8\% \\ 
& instruments & 65 & \textit{harmonica - harp} & 96.9\% &90.8\% & 96.9\% & 67.7\% & 96.9\% \\ 
& planets & 60 & \textit{Mars - Venus} & 31.7\% & 3.3\%& 21.7\% & 100.0\% & 5.0\% \\ 
\midrule
Ent. & synonyms & 894 & \textit{happy - joyful} & 97.5\% & 99.7\% & 86.1\% & 70.5\% & 92.1\% \\ 
\midrule
& total & 8,193 &  & 51.9\% &65.6\% & 62.2\% & 85.8\% & 83.5\% \\ 
\bottomrule
\end{tabular}
\vspace*{-10pt}
\caption{The number of instances and accuracy per category achieved by each model.}
\label{tab:accuracy_by_category}
\vspace*{-12pt}
\end{table*}
 
\subsection{Experimental Settings}
\label{sec:experimental_settings}

We trained each model on 3 different datasets: (1) SNLI train set, (2) a union of the SNLI train set and the MultiNLI train set, and (3) a union of the SNLI train set and the SciTail train set. The motivation is that while SNLI might lack the training data needed to learn the required lexical knowledge, it may be available in the other datasets, which are presumably richer. 

\subsection{Results}
\label{sec:results}

Table~\ref{tab:results} displays the results for all the models on the original SNLI test set and the new test set. Despite the task being considerably simpler, the drop in performance is substantial, ranging from 11 to 33 points in accuracy. Adding MultiNLI to the training data somewhat mitigates this drop in accuracy, thanks to almost doubling the amount of training data. We note that adding SciTail to the training data did not similarly improve the performance; we conjecture that this stems from the differences between the datasets. 

\textsc{Kim} substantially outperforms the other neural models, demonstrating that lexical knowledge is the only requirement for good performance on the new test set, and stressing the inability of the other models to learn it. Both WordNet-informed models leave room for improvement: possibly due to limited WordNet coverage and the implications of applying lexical inferences within context. 

\section{Analysis}
\label{sec:analysis}
We take a deeper look into the predictions of the models that don't employ external knowledge, focusing on the models trained on SNLI. 

\subsection{Accuracy by Category}
\label{sec:accuracy_by_category}

Table~\ref{tab:accuracy_by_category} displays the accuracy of each model per replacement-word category. The neural models tend to perform well on categories which are frequent in the training set, such as \textit{colors}, and badly on categories such as \textit{planets}, which rarely occur in SNLI. These models perform better than the WordNet baseline on entailment examples (\textit{synonyms}), suggesting that they do so due to high lexical overlap between the premise and the hypothesis rather than recognizing synonymy. We therefore focus the rest of the discussion on contradiction examples. 

\subsection{Accuracy by Word Similarity}
\label{sec:accuracy_by_cosine}

The accuracies for \textit{ordinals}, \textit{nationalities} and \textit{countries} are especially low. We conjecture that this stems from the proximity of the contradicting words in the embedding space. Indeed, the Decomposable Attention model---which does not update its embeddings during training---seems to suffer the most. 

Grouping its prediction accuracy by the cosine similarity between the contradicting words reveals a clear trend that the model errs more on contradicting pairs with similar pre-trained vectors:\footnote{We ignore multi-word replacements in \S\ref{sec:accuracy_by_cosine} and \S\ref{sec:accuracy_by_frequency}.}

\vspace{-5pt}
\setlength\tabcolsep{4pt}
\begin{table}[!h]
\small
\centering
\begin{tabular}{|l|c|c|c|c|c|}
\hline
Similarity & 0.5-0.6 & 0.6-0.7 & 0.7-0.8 & 0.8-0.9 & 0.9-1.0 \\
\hline
Accuracy & 46.2\% & 42.3\% & 37.5\% & 29.7\% & 20.2\% \\
\hline
\end{tabular}
\vspace{-5pt}
\label{tab:acc_by_cos_sim}
\vspace{-5pt}
\end{table} 

\subsection{Accuracy by Frequency in Training}
\label{sec:accuracy_by_frequency}

Models that fine-tune the word embeddings may benefit from training examples consisting of test replacement pairs. Namely, for a given replacement pair ($w_p$, $w_h$), if many training examples labeled as contradiction contain $w_p$ in the premise and $w_h$ in the hypothesis, the model may update their embeddings to optimize predicting contradiction. Indeed, we show that the ESIM accuracy on test pairs increases with the frequency in which their replacement words appear in contradiction examples in the training data: 

\vspace{-5pt}
\setlength\tabcolsep{3pt}
\begin{table}[!h]
\small
\centering
\begin{tabular}{|l|c|c|c|c|c|c|}
\hline
Frequency & 0 & 1-4 & 5-9 & 10-49 & 50-99 & 100+ \\
\hline
Accuracy & 40.2\% & 70.6\% & 91.4\% & 92.1\% & 97.5\% & 98.5\% \\
\hline
\end{tabular}
\vspace{-5pt}
\label{tab:acc_by_freq_bin}
\vspace{-5pt}
\end{table}

This demonstrates that the model is capable of learning lexical knowledge when sufficient training data is given, but relying on explicit training examples is a very inefficient way of obtaining simple lexical knowledge.

\section{Conclusion}
\label{sec:conclusion}

We created a new NLI test set with the goal of evaluating systems' ability to make inferences that require simple lexical knowledge. Although the test set is constructed to be much simpler than SNLI, and does not introduce new vocabulary, the state-of-the-art systems perform poorly on it, suggesting that they are limited in their generalization ability. The test set can be used in the future to assess the lexical inference abilities of NLI systems and to tease apart the performance of otherwise very similarly-performing systems. 

\section*{Acknowledgments}

We would like to thank Qian Chen for evaluating KIM on our test set. This work was supported in part by the German Research Foundation through the German-Israeli Project Cooperation (DIP, grant DA 1600/1-1), an Intel ICRI-CI grant, Theo Hoffenberg, and the Israel Science Foundation grants 1951/17 and 1555/15. Vered is also supported by the Clore Scholars Programme (2017), and the AI2 Key Scientific Challenges Program (2017).

\bibliography{bib}
\bibliographystyle{acl_natbib}

\end{document}